УДК 004.83

**Pogorilyy S.D., Kramov A.A.**

# Method of noun phrase detection in Ukrainian texts

**Introduction.** The area of natural language processing considers AI-complete tasks that cannot be solved using traditional algorithmic actions. Such tasks are commonly implemented with the usage of machine learning methodology and means of computer linguistics. One of the preprocessing tasks of a text is the search of noun phrases. The accuracy of this task has implications for the effectiveness of many other tasks in the area of natural language processing. In spite of the active development of research in the area of natural language processing, the investigation of the search for noun phrases within Ukrainian texts are still at an early stage.

**Purpose**. Comparative analysis of the main methods of noun phrases detection in English and Ukrainian texts. The creation of a complex method for the detection of noun phrases in texts according to the features of the Ukrainian language. The performing of experimental examination of the suggested method on the corpus of Ukrainian articles.

**Results**. The different methods of noun phrases detection have been analyzed. The expediency of the representation of sentences as a tree structure has been justified. The key disadvantage of many methods of noun phrase detection is the severe dependence of the effectiveness of their detection from the features of a certain language. Taking into account the unified format of sentence processing and the availability of the trained model for the building of sentence trees for Ukrainian texts, the Universal Dependency model has been chosen. The complex method of noun phrases detection in Ukrainian texts utilizing Universal Dependencies means and named-entity recognition model has been suggested. Experimental verification of the effectiveness of the suggested method on the corpus of Ukrainian news has been performed. Different metrics of method accuracy have been calculated.

**Conclusions**. The results obtained can indicate that the suggested method can be used to find noun phrases in Ukrainian texts. An accuracy increase of the method can be made with the usage of appropriate named-entity recognition models according to a subject area.

\_ \_ \_ \_ \_

Здійснено порівняльний аналіз основних автоматизованих методів пошуку іменних груп та іменованих сутностей в англомовних та україномовних текстах; обґрунтовано доцільність використання моделі Universal Dependencies. Запропоновано комплексний метод на основі аналізу деревовидної синтаксичної структури речення та моделі виявлення іменованих сутностей. Здійснено експериментальну перевірку ефективності пропонованого методу та показано доцільність його використання для пошуку іменних груп в україномовних текстах.

Погорілий С.Д., Крамов А.А

## Метод виявлення іменних груп в україномовних текстах

### Вступ

Постійна динаміка росту потужностей обчислювальних систем обумовлює використання методів машинного навчання для формалізації та вирішення задач, подібних до дій людини. Задачі такого типу, що не можуть бути вирішені за допомогою алгоритмічних дій, називають *AI-повними*. Зважаючи на постійне зростання обсягу текстової інформації, актуальною проблемою є автоматизований аналіз природної мови для отримання структурованих даних: розпізнавання мовлення, машинний переклад, розв'язання лексичних неоднозначностей тощо. Зазначені задачі варто відносити до завдань комп'ютерної лінгвістики та методології машинного навчання, а саме до галузі обробки природної мови (англ. *natural language processing – NLP*). Незважаючи на відмінність поставлених цілей, задачі NLP містять спільний початковий етап, а саме *попередню обробку вхідних даних* (текстової інформації). Попередня обробка тексту необхідна для формального представлення текстової інформації у вигляді структурованих даних. Засоби формалізації тексту можуть відрізнятися відповідно до поставленої задачі, однак варто виділити наступні кроки попередньої обробки тексту, які використовуються в більшості задач обробки природної мови:

- токенізація (англ. *tokenization*) – процес розбиття тексту на речення, а речення на окремі слова;
- розмічування слів – співставлення кожній атомарній одиниці тексту (слову) частини мови, роду, відмінку;
- лематизація – приведення слова до нормальної форми; наприклад, для української мови нормальною формою іменників є його представлення у називному відмінку, а дієслово трансформується до інфінітивної форми;
- *пошук сутностей (іменних груп) в тексті*.

На відміну від попередніх кроків, які здійснюються шляхом використання заздалегідь визначених правил і різнотипних словників, останній крок потребує детальнішого аналізу. Зважаючи на постійну зміну лексичного складу мови, виявлення сутностей потребує спільного використання методології машинного навчання та засобів комп'ютерної лінгвістики. Таким чином, пошук іменних груп в тексті варто відносити до класу AI-повних задач, що не можуть бути формалізовані визначеним алгоритмом.

Отже, завдання пошуку іменних груп є важливим етапом в процесі вирішення інших задач обробки природної мови. Підвищення точності детектування іменних груп в тексті дозволить в подальшому покращити ефективність застосування методів вирішення задач, залежних від цього пошуку. Наявність актуальних робіт щодо визначення іменних груп в різних мовах свідчить про важливість дослідження методів вирішення цієї задачі. Незважаючи на активний розвиток досліджень в напрямку обробки природних мов, дослідження пошуку іменних груп для україномовних текстів знаходиться на початковому етапі.

**Мета роботи**:

- аналіз існуючих методів автоматизованого пошуку іменних груп в англомовних та україномовних текстах;

- створення комплексного методу детектування іменних груп на основі дерева залежностей речення і моделі розпізнавання іменованих сутностей;
- проведення експериментальної перевірки зазначеного методу для корпусу текстів української мови.

## Концепт іменної групи

Термін *«іменна група»* запозичений з англомовного варіанту *noun phrase*. В українській мові цей термін трактується як іменникове (субстантивне) словосполучення – словосполучення з іменником у ролі головного слова [1]. Однак особовий займенник (я, ти, він), який вказує на конкретний об'єкт, також може використовуватися як окрема сутність в реченні, тому надалі будемо розглядати термін «іменна група» як іменникове чи займенникове словосполучення.

Розглянемо більш детально варіанти формування іменної групи в українській мові.

Іменник у ролі головного слова може сполучатися:

- з прикметником (*червоний колір, смачний обід*);
- з іменником у непрямих відмінках з прийменником або без нього (*брат Петра, думки про майбутнє*);
- з займенником (*ця думка, моя мрія*);
- з дієприкметником (*зів'ялі квіти, пожовкла трава*);
- з прислівником (*читання вголос*);
- з числівником (*два кольори*);
- з інфінітивом (*бажання вчитися*).

Займенник у ролі головного слова може сполучатися:

- з іменником (*хтось зі звірів, когось із тварин*);
- з прикметником (*щось цікаве*);
- з займенником (*кожного з нас*).

Незважаючи на наявність вказаних вище правил, процес пошуку іменних груп не є тривіальним для української мови. Для мов, в яких існує клас артиклів (наприклад, англійська), індикатором іменної групи є *детермінатив* – словоформа чи морфема, яка супроводжує іменну групу та узагальнює інформацію про групу (рід, число тощо). Наприклад, у синтаксичних теоріях англійської мови вважається, що будь-яка іменна група містить детермінатив [2]. Українська мова відноситься до класу мов без артиклів. Наразі немає однозначної відповіді щодо наявності в цьому класі детермінативу в іменних групах. Питання пошуку іменної групи в ієрархічній структурі для мов без артиклів розглянуто в роботі [3] на прикладі російської мови. В цій роботі розглядаються загальні принципи узгодження головного слова іменної групи з залежними словами. Зокрема, доводиться *ієрархічна побудова іменної групи в російській мові* та проводиться аналіз узгодження слів іменної групи по числу та роду.

Алгоритм детектування головного слова та дочірніх слів повинен враховувати принаймні наступні особливості текстової інформації:

- відсутність артиклів в українській мові, які певною мірою ідентифікують іменні групи в іноземних мовах;
- неструктурована побудова речення (можливий зворотній порядок слів);
- наявність фразеологізмів, власних назв та слів іншомовного походження.

**Порівняльний аналіз існуючих методів пошуку іменних груп**

Проблема пошуку іменних груп в тексті активно вирішується для англомовних текстів, про що свідчить наявність робіт [4-6]. Метод $n$-грам [4] полягає в пошуку всіх послідовностей слів, які зустрічаються в тексті, довжиною $k$ ($1 \leq k \leq n$); послідовність повинна знаходитися в межах одного речення. Такий підхід ефективно використовується для отримання ознак у задачі класифікації текстів, але з точки зору семантичного значення групи метод $n$-грам має значний недолік: фіксований розмір послідовностей. Фіксований розмір групи призводить до втрати смислового навантаження набору слів, які входять до послідовності. Наприклад, іменна група «*Міністерство освіти і науки України*» може інтерпретуватися як послідовність «*Міністерство освіти і*», яка не відображає її семантичний зміст користувачу. Принцип роботи методу NPFST [5] полягає у використанні заздалегідь описаних шаблонів іменних груп. Шаблони представлені у вигляді рядків – регулярних виразів, в яких змінні елементи відображають різні частини мови. Нижче наведений приклад такого регулярного виразу:

$$(A\,|\,N)*N(PD*(A\,|\,N)*N)* \tag{1}$$

Після здійснення операцій токенізації та розмічування слів, кожному слову ставиться у відповідність текстова мітка частини мови ($A, N, P$ тощо). Кожне слово замінюється на потрібну мітку, тобто речення тексту трансформуються у рядки, що містять не слова, а мітки. Далі до отриманого тексту застосовується набір шаблонів. У випадку детектування відповідності шаблону частині тексту виконується екстракція знайденої частини з подальшим зворотнім перетворенням від мітки до слова. Недоліком такого підходу є залежність набору шаблонів від особливостей мови та стилістики тексту. Крім того, цей метод не є масштабованим, адже постійне збільшення кількості шаблонів підвищує ймовірність колізії регулярних виразів, що призведе до некоректної роботи методу.

В 2016 році був запропонований універсальний підхід (*Universal Dependencies – UD*) до перетворення текстової інформації в деревовидну структуру [6]. Універсальність підходу передбачає узагальнення різнотипних зв'язків між словами речення ***незалежно від мови***. В роботі запропонована загальна схема впорядкування слів речення залежно від їх частини мови: іменник і прикметник, іменник і займенник тощо. Внаслідок такої уніфікації формату для різних мов, за допомогою зусиль відкритої спільноти з різних країн вдалося створити набір моделей перетворення вхідного тексту в деревовидну структуру. Поточна версія UD 2.3 містить підтримку 76 мов. Для україномовних текстів також були підготовлені вхідні дані (розмічені тексти) і в подальшому була навчена відповідна модель [7]. Приклад такої структури наведено на рис. 1.

Іменник чи займенник, що є вершиною дерева та містить дочірні вузли, можна трактувати як потенційне головне слово своєї групи. Обхід дерева дозволяє поставити у відповідність кожному потенційному головному слову іменної групи набір залежних слів, причому таке зіставлення може відбуватися і для глибших рівнів у рекурсивний спосіб.

Враховуючи належність української та російської мов до спільного класу мов без артиклів та ієрархічну структуру іменних груп в російськомовних текстах, доцільним є здійснення пошуку іменних груп в україномовних текстах за допомогою аналізу моделі UD.

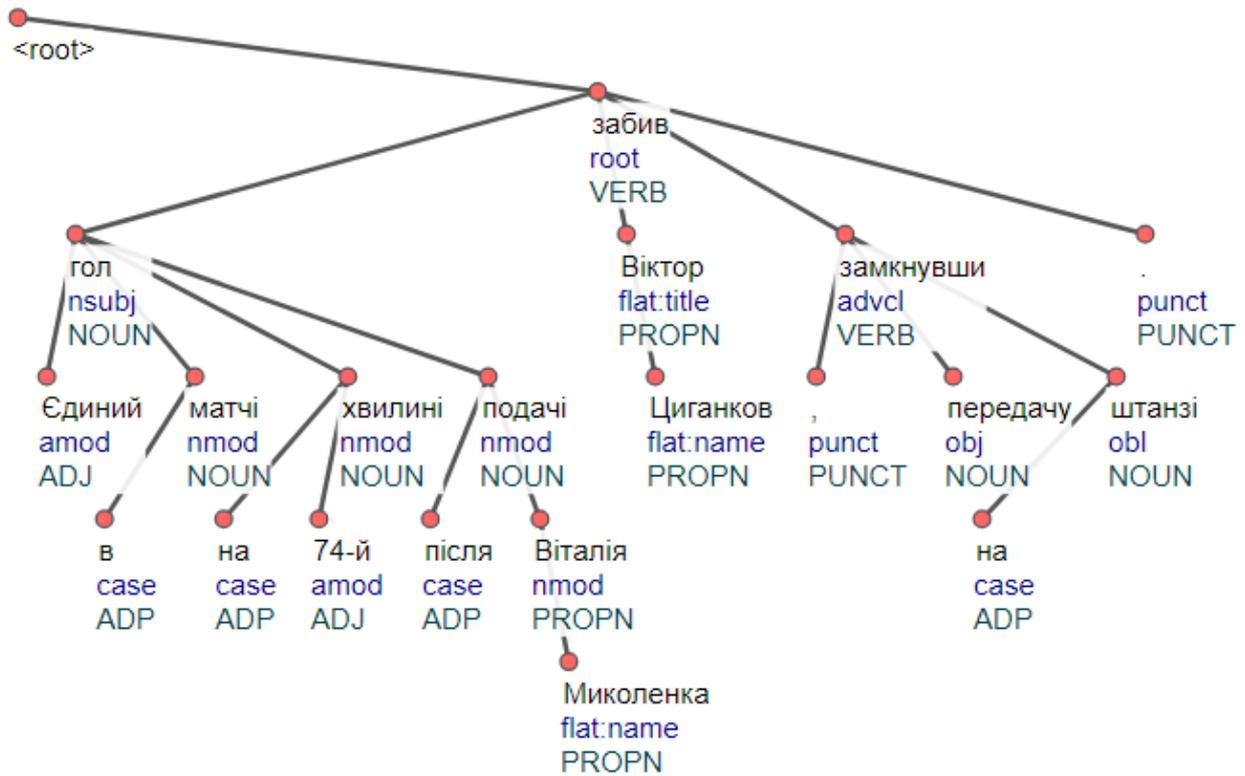

*Рис. 1 – Приклад представлення тексту в деревовидній структурі*

**Методи розпізнавання іменованих сутностей в тексті**

Окремо варто розглянути питання виділення іменованих сутностей в тексті. Принцип встановлення порядку та узгодження слів в іменованій сутності може дещо відрізнятися від результату аналізу деревовидної структури речення. Така відмінність може виникнути в результаті унікальної структури іменованої сутності, що не підпорядковується загальним правилам побудови іменної групи, та некоректного перетворення вхідного тексту в деревовидну структуру. Наприклад, розглянемо наступне речення: «**Група акціонерів компанії Facebook (1)** наполягає на тому, що **засновник соціальної мережі Марк Цукерберг (2)** повинен втратити **посаду голови правління (3)**.». Напівжирним шрифтом з підкресленням виділені іменні групи речення, а в дужках вказаний порядковий номер групи. Результат перетворення речення в деревовидну структуру зображено на рис. 2. Групи (1) і (3) можуть бути ідентифіковані коректно, адже їхні елементи розташовані у рекурсивний спосіб відповідно до очікуваної структури цих груп. Розглянемо групу (2). Батьківським словом відповідної групи в дереві є слово «засновник», яке помічене як іменник, тобто може бути головним словом групи. Виконавши обхід дочірніх елементів у рекурсивний спосіб, отримуємо наступну послідовність: «засновник соціальної мережі Марк». Порівнюючи з очікуваним результатом, відсутнє слово «Цукерберг». Розглянувши детальніше відповідну область дерева, можна побачити, що слово «Цукерберг» не потрапляє до списку дочірніх елементів слова «засновник»; крім того, частина мови цього слова ідентифікована як дієслово.

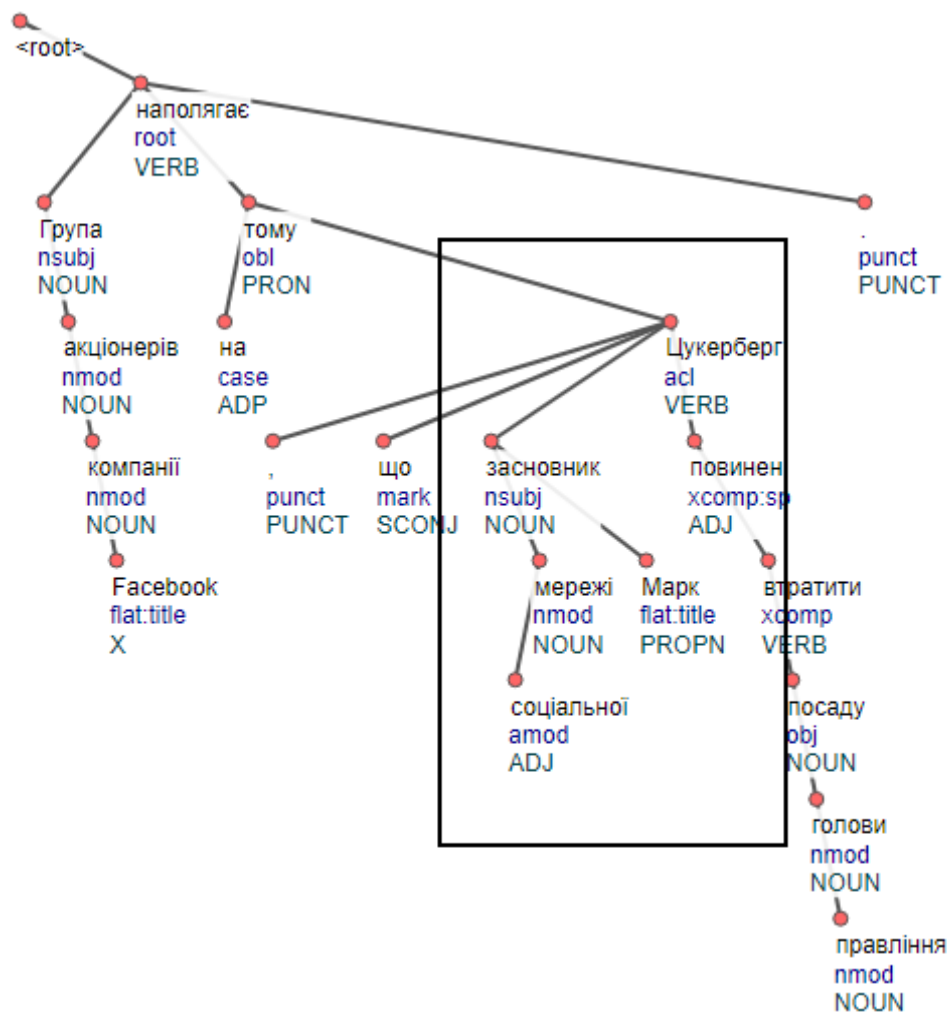

*Рис. 2 – Деревовидна структура речення з некоректним впорядкуванням елементів. Чорною лінією обмежена область дерева з помилковим розташуванням вершин дерева.*

Некоректне розмічування слів частинами мови може траплятися через відсутність слова в морфологічному словнику; найчастіше така ситуація може виникати для власних назв. Фрагменти тексту з власними назвами можливо знайти за допомогою додаткового використання моделі виділення іменованих сутностей (англ. named-entity recognition – NER). У наведеному вище прикладі, модель виділення іменованих сутностей може ідентифікувати пару слів «Марк Цукерберг» як особу. Подальше об'єднання множин «засновник соціальної мережі Марк» і «Марк Цукерберг», які мають спільні елементи (слово «Марк»), призводить до отримання очікуваного результату: «засновник соціальної мережі Марк Цукерберг». Отже, використання моделі виділення іменованих сутностей може використовуватися як додатковий інструмент під час пошуку іменних груп в тексті. Ефективність застосування цієї моделі залежить від типу об'єктів, які вона здатна розпізнавати, та предметної області тексту, що аналізується.

Розглянемо існуючі рішення та пропозиції щодо виділення іменованих сутностей в україномовних текстах. Відкритою спільнотою фахівців lang-uk було здійснено навчання моделі NER на проанотованому корпусі української мови [8]. Для пошуку іменованих сутностей використовувалася відкрита бібліотека MITIE, яка має інтерфейси для багатьох мов програмування: C/C++, Python, Java. Тренування моделі здійснювалося на вибірці з 229 текстів для розпізнавання сутностей, які належать до наступних категорій:

- персона;
- локація;
- організація;
- різне.

Також варто виділити роботу [9], в якій пропонується використовувати підхід пошуку сутностей за шаблонами. Для кожного типу сутності створюється окремий набір правил, який дозволяє однозначно його ідентифікувати. Алгоритм виділення сутностей використовує GLR-парсер. В цій роботі виділення іменованих сутностей здійснювалося для наступних категорій:

- персона;
- організація;
- географічний об'єкт.

Для порівняння ефективності розглянутих методів доцільно розглянути їхні значення F-міри [10]. F-міра ($F$) – це середнє гармонійне значення точності ($Precision$) і повноти ($Recall$):

$$Precision = \frac{TP}{TP + FP} \qquad (2)$$

$$Recall = \frac{TP}{TP + FN} \qquad (3)$$

$$F = 2\frac{Precision \times Recall}{Precision + Recall} \qquad (4)$$

де *TP* – кількість коректно розпізнаних сутностей; *FP* – кількість сутностей, які не були розпізнані; *FN* – кількість сутностей, які були ідентифіковані моделлю, але відсутні в експертній розмітці тексту. Значення F-міри досягає значення 0.8 для моделі, створеної спільнотою lang-uk; 0.54 для моделі на основні пошуку шаблонів. Незважаючи на зазначений показник F-міри, модель на основі пошуку шаблонів може використовуватися як додатковий інструмент пошуку іменованих сутностей в текстах певної предметної області.

### Метод пошуку іменних груп в україномовних текстах

Пошук іменних груп в україномовних текстах пропонується здійснювати за рахунок аналізу деревовидної структури речення, отриманої за допомогою підходу Universal Dependencies (UD) [11]. Враховуючи, що головним словом іменної групи може бути іменник чи займенник, спочатку розглянемо вершини з відповідними частинами мови. Відповідно до категорій розмітки тексту частинами мови, до потенційних головних слів іменної групи варто віднести слова з наступними категоріями:

– *NOUN* (іменник);

– *PRON* (займенник);

– *PROPN* (власна назва);

– *X* (інша частина мови).

З наведеного списку варто виділити два пункти: *PROPN* і *X*. Власна назва (ім'я, прізвище, місто тощо) може бути розмічена як категорія *PROPN*, яка теж здатна формувати іменну групу або входити до складу існуючої. Категорія *X* встановлюється слову у випадку, коли модель не може передбачити частину мови. Однак, слова з такою категорією можуть мати додатковий параметр *Foreign=Yes*, який вказує, що це слово іншомовного походження. Проаналізувавши 2500 різних текстів, написаних українською мовою, було виявлено, що 99% слів іншомовного походження є сутностями, які вказують на певний об'єкт. Таким чином, доцільно додатково розглядати слова з тегом *X* і додатковим параметром *Foreign=Yes* як потенційне головне слово групи чи складову частину іншої групи.

Визначивши тип вершин, які можуть розглядатися як головне слово іменної групи, визначимо правила приєднання дочірніх слів до іменної групи батьківського слова. Відповідно до деревовидної структури речення, можна зробити припущення, що всі дочірні вершини головного слова логічно пов'язані з ним та входять до його іменної групи. Однак таке припущення є хибним, враховуючи наступні фактори:

– похибка попередньої обробки тексту, а саме процесу токенізації та розмічення слів частинами мови;
– похибка власне моделі побудови деревовидної структури;
– граматичні та пунктуаційні помилки у вхідному тексті.

Отже, потрібно сформувати набір правил приєднання дочірніх вершин до батьківської іменної групи. Розглянемо окремо принципи входження потенційних головних слів, дієслів та інших дочірніх елементів до поточної батьківської групи.

**Загальний підхід приєднання дочірнього елементу до іменної групи**

Враховуючи правила формування іменних груп (субстантивних словосполучень) в українській мови, до складу іменної групи можуть входити слова з наступними частинами мови (в дужках вказуються відповідні теги моделі UD): прикметник (*ADJ*), прислівник (*ADV*), прийменник (*ADP, DET, AUX*), числівник (*NUM*), іменник (*NOUN, PROPN, X*), займенник (*PRON*), дієслово (*VERB*), знаки пунктуації (*PUNCT*). Слова, розмічені як інші частини мови, чи додаткові символи (знаки арифметичних операцій, сполучники тощо) не включаються до іменної групи. Крім того, всі елементи групи повинні розташовуватися в тексті послідовно. Якщо між дочірнім елементом, який може входити до групи, знаходиться заборонений елемент, входження зазначеного дочірнього елементу до групи відхиляється. Додатковою перевіркою умови приєднання дочірнього елементу до групи може бути уточнення його узгодженості з головним словом за числом і родом. Наприклад, у такий спосіб узгоджені іменні групи «дві медалі» (за числом) і «кваліфікований фахівець» (за родом). В українській мові є різнотипні варіанти такого узгодження, пов'язані з граматичною складовою мови. Для прикладу розглянемо наступне словосполучення: «п'ятдесят один кілометр». Головне слово «кілометр» має число однини, хоча в цьому контексті складений числівник «п'ятдесят один» вказує на число множини. Використовується наступне правило: після числівника «один», навіть якщо він входить до складених числівників, іменник вживається в формі однини. Застосування набору правил узгодження головного слова з підрядними дозволяє перевірити можливість їх приєднання до групи. Однак необхідною умовою використання згаданого вище набору правил є врахування всіх аспектів формування словосполучення, що для веб-ресурсів більшості сучасних ЗМІ є малоймовірним. Помилкові вирази «два з половиною місяц**я**», «заступни**ця** Міністра культури» можуть зустрічатися в новинних текстах чи розмовній мові. Таким чином, було вирішено не здійснювати перевірку узгодження слів іменної групи за числом і родом для коректної обробки текстів з різною стилістикою.

**Приєднання дієслова до іменної групи**

Іменна група з іменником у ролі головного слова може містити дієслово у формі інфінітиву (наприклад, «бажання вчитися» чи «необхідність працювати»). Отже, необхідно уточнити, чи дочірнє слово (дієслово) знаходиться у формі інфінітиву. Для цього можна скористатися додатковим параметром моделі UD *VerbForm*. Параметр *VerbForm* присутній тільки для дієслів; у випадку представлення дієслова в формі інфінітиву, атрибут набуває значення *Inf*.

**Приєднання потенційного головного слова до іменної групи**

Найскладнішим є прийняття рішення щодо входження дочірнього потенційного головного слова (ДПГС) до батьківської іменної групи, адже дочірній елемент може формувати окрему групу. Проаналізувавши деревовидні структури україномовних текстів та відповідні синтаксичні зв'язки моделі UD [12], було вирішено сформувати наступні критерії входження ДПГС до іменної групи:

– наявність відповідного типу синтаксичного зв'язку;
– відсутність заборонених елементів серед дочірніх вершин ДПГС.

Розглянемо типи синтаксичного зв'язку між ДПГС та батьківською вершиною, необхідні для входження ДПГС до іменної групи. До таких типів належать:

– *flat* – встановлюється між словами, які входять до складу власних назв чи дат, тобто у випадку, коли невідома внутрішня синтаксична структура виразу;

– *nmod* – зв'язок між елементами, один з яких модифікує інший; зазвичай, такий зв'язок передбачає представлення дочірнього елемента в родовому відмінку.

Наведені типи зв'язку можна відслідкувати у реченні: «Під час позачергових парламентських виборів 2014 року, майбутній міністр Лілія Гриневич потрапила до парламенту». На рис. 3 зображена деревовидна структура цього речення.

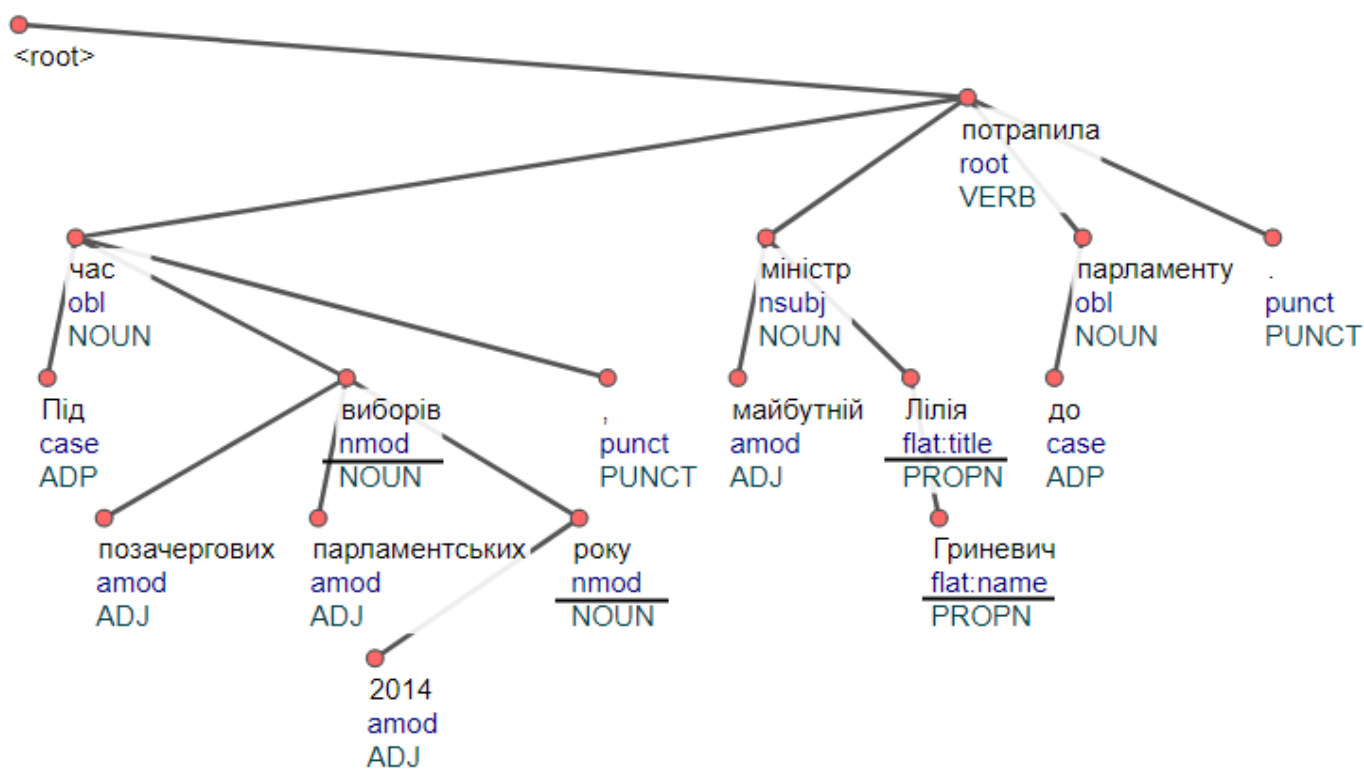

*Рис. 3 – Деревовидна структура речення, що містить іменні групи, утворені зв'язками flat і nmod*

На наведеному вище прикладі, зв'язок *flat* дозволяє з'єднати власні іменники «Лілія» і «Гриневич» зі словом «міністр», у такий спосіб утворюючи іменну групу «майбутній міністр Лілія Гриневич». Зв'язок *nmod* простежується в першій частині реченням між словами «час» і «виборів», «виборів» і «року». Поєднуючи всі відповідні слова у рекурсивний спосіб відповідно до їх порядку розташування в тексті, отримуємо іменну групу «під час позачергових парламентських виборів 2016 року».

Варто згадати ще два типи зв'язків, які разом зі *flat* відносяться до типу MWE (multiword expressions – багатослівні вирази): *fixed* і *compound*. Зв'язок *fixed* вказує на стійке словосполучення: «до того ж»,

«мало не сто років» тощо. Щодо *compound,* цей тип зв'язку зазвичай використовується для композицій з числами. Вказані зв'язки можуть використовуватися для формування різних структурних одиниць в тексті, але їх застосування не є доцільним для відстеження відношень в іменних групах.

**Обхід деревовидної структури речення**

Розглянувши критерії відбору потенційного головного слова групи та правила приєднання дочірніх елементів до батьківської групи, варто звернути увагу на порядок обходу деревовидної структури. Зрозуміло, що обхід структури такого типу здійснюється у рекурсивний спосіб (використовується центрований порядок). Необхідно зазначити, що елементи іменної групи розташовуються в тексті послідовно, тобто між цими елементами немає сторонніх слів, які не відносяться до групи. Таким чином, у випадку перевірки входження дочірнього елементу до батьківської групи потрібно здійснювати додатковий аналіз того, чи входять до групи елементи, які розташовані в реченні між поточним дочірнім елементом та батьківською вершиною. Для уникнення зазначених перевірок пропонується здійснювати обхід дочірніх вершин у наступний спосіб:

– від найближчого дочірнього елементу, розташованого ліворуч від головного слова в тексті, до крайнього лівого дочірнього елементу;
– від найближчого дочірнього елементу, розташованого праворуч від головного слова в тексті, до крайнього правого дочірнього елементу.

Такий порядок обходу дочірніх вершин дозволяє уникнути зазначеної вище додаткової перевірки, адже у випадку виявлення несумісності дочірнього елементу з батьківською вершиною, всі наступні вершини можуть розглядатися як об'єкти, незалежні від головного слова. Приклад порядку обходу деревовидної структури речення зображено на рис. 4.

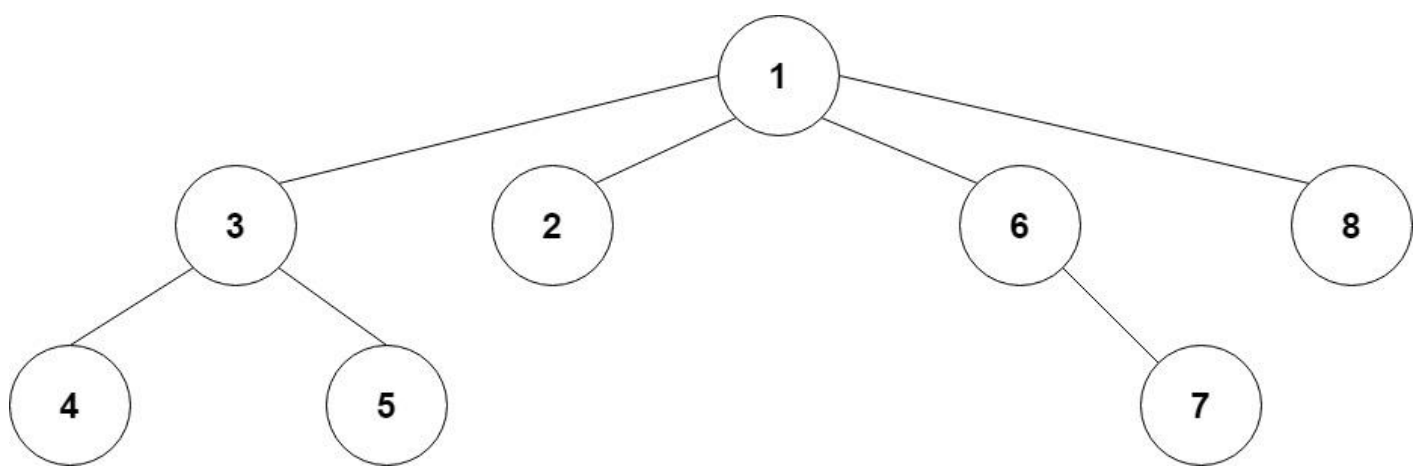

*Рис. 4 – Приклад порядку обходу деревовидної структури речення*

**Пошук іменованих сутностей в тексті**

Додатково до пропонованого пошуку іменних груп в тексті варто застосовувати виділення іменованих сутностей. Як було зазначено раніше, виділення іменованих сутностей дозволяє виявити сполучення слів, які неможливо ідентифікувати за допомогою аналізу отриманої деревовидної структури (через некоректне розмічування слів чи похибки роботи моделі синтаксичного розбору речення). Пошук іменованих сутностей варто розпочинати із застосування *газетирів* – словників, які містять перелік географічних назв з додатковою інформацією про них. З точки зору автоматизованої обробки тексту, під терміном «газетир» зазвичай розглядається список власних назв відповідно до предметної області дослідження. Результатом застосування газетирів до вхідного тексту є набір груп – іменованих сутностей, кожна з яких містить індекси-вказівники на певні слова тексту. Для формування газетиру були використані наступні бази даних:

- перелік найпопулярніших прізвищ та імен жінок і чоловіків (форма «прізвище ім'я»);
- перелік країн;
- перелік міст.

Після отримання результату застосування газетиру до тексту, наступним кроком є запуск навченої моделі виділення іменованих сутностей. Як модель виділення іменованих сутностей було обрано NER-модель спільноти lang-uk; для застосування моделі було використано відкриту бібліотеку

MITIE. Вихідним результатом роботи моделі є набір об'єктів, кожен з яких відповідає розпізнаній іменованій сутності та має наступні атрибути:

– діапазон індексів слів, які входять до іменованої сутності;
– категорія іменованої сутності;
– оцінка «впевненості» моделі в тому, що поточна іменована сутність розпізнана коректно.

Варто звернути увагу на останній атрибут. Відповідно до документації бібліотеки, чим більше значення оцінки «впевненості», тим вища ймовірність коректного передбачення. Зважаючи на відсутність еталонного порогового значення зазначеної оцінки, було вирішено встановити це значення експериментальним шляхом за допомогою розрахунку F-міри моделі на множині україномовних текстів. Відповідно до розміченої тестової вибірки текстів, отримане оптимальне порогове значення оцінки «впевненості» моделі рівне 0.8. Вказане значення оцінки моделі було використано в подальших експериментальних перевірках цієї роботи.

## Експериментальна перевірка методу

Для здійснення експериментальної перевірки ефективності пропонованого методу було створено відповідне застосування; серверна мова програмування – Python 3.6. Відповідно до послідовних етапів здійснення перевірки роботи методу, застосування складається з трьох компонентів:

– веб-сторінка розмітки іменних груп в тексті;
– модуль пошуку іменних груп в україномовних текстах;
– утиліта розрахунку оцінки ефективності роботи методу.

### Веб-сторінка розмітки іменних груп

Оцінка F-міри методу можлива за наявності попередньо розміченого тексту – комбінацій слів та символів, які експерт позначив як іменні групи. Перевірочну вибірку текстів було сформовано зі статей новинних порталів різної тематики. Протягом дослідження оброблено 100 різних документів; загальна кількість знайдених іменних груп: 1488. Для формування перевірочної вибірки було створено веб-сторінку, яка здійснює графічне відображення результату токенізації вхідного україномовного тексту та дозволяє виконувати групування слів та символів в іменні групи. Приклад обробки вхідного тексту за допомогою зазначеної веб-сторінки зображено на рис. 5.

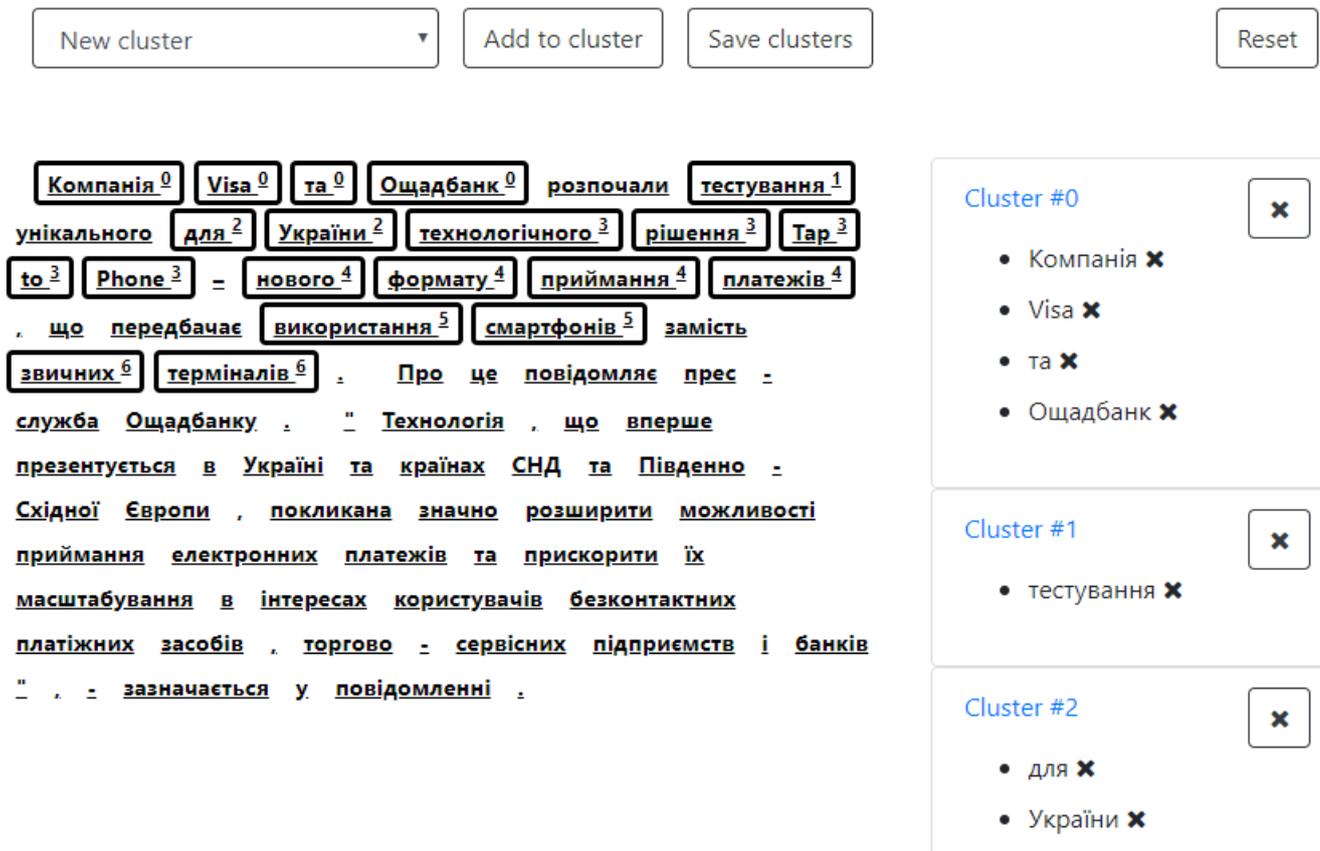

*Рис. 5 – Веб-інтерфейс розмітки іменних груп в тексті; розпізнані об'єкти підкреслені лінією, іменні групи, позначені користувачем, додаткові виділені суцільною рамкою; індекс у правому верхньому куті об'єкту вказує на номер іменної групи, до якої він входить.*

Принцип роботи веб-сторінки наступний: користувач копіює україномовний текст із зовнішнього джерела та вставляє його в текстове поле сторінки. Далі він натискає на кнопку «Recognize», після чого виконується токенізація вхідного тексту. Користувач вибирає об'єкти, які відносяться до спільної іменної групи, та формує відповідний кластер. Після закінчення процесу розмітки тексту користувач натискає на кнопку «Save clusters», зберігаючи створені кластери в базі даних. Для створення інтерактивного режиму формування іменних груп використано фреймворк Marionette.js. Збереження сформованих кластерів для подальшої оцінки ефективності роботи методу здійснено за допомогою реляційної бази даних MySQL.

**Модуль пошуку іменних груп**

Модуль пошуку іменних груп в україномовних текстах реалізовано мовою програмування Python. Створений модуль розміщено на платформі The Python Package Index (PyPI), що дозволяє виконувати імпорт модулю у сторонні проекти. Інструкції щодо встановлення та використання модулю доступні за посиланням [13]. Модуль містить сторонні пакети, які необхідно попередньо встановити для коректної роботи модулю. Передбачено додаткове підключення моделі пошуку іменованих сутностей в україномовному тексті, а також використання сторонніх газетирів.

**Результати оцінки ефективності роботи методу**

Метрикою оцінки ефективності роботи методу було обрано три параметри: точність, повнота та F-міра. Здійснення розрахунку параметрів виконано для двох режимів: повної і часткової відповідностей. Для детектування повної відповідності необхідно, щоб прогнозована та розмічена іменні групи повністю співпадали (порівняння за символами та позиціями в тексті); іменні групи вважаються частково відповідними одна одній, якщо хоча б одна межа груп співпадає (початкове чи кінцеве слово). Крім того, було вирішено розрахувати метрику для трьох різних варіантів використання моделей:

– з моделями аналізу деревовидної структури речення і пошуку іменованих сутностей (UD+NER);

– з моделлю аналізу деревовидної структури речення, але без використання моделі пошуку іменованих сутностей (UD);

– без використання моделей, зазначених в попередніх пунктах; в цьому випадку окремими іменними групами вважаються іменники та особові займенники (–).

| Метрика<br>Варіанти моделей | Точність | Повнота | F-міра |
|---|---|---|---|
| UD+NER | 0.552 | 0.573 | 0.559 |
| UD | 0.555 | 0.572 | 0.560 |
| – | 0.175 | 0.039 | 0.062 |

*Табл. 1 – Оцінка ефективності роботи методу з різними варіантами моделей для режиму повної відповідності*

В табл. 1 наведено оцінки ефективності роботи різних варіантів використання моделей відповідно до розрахованої метрики для режиму повної відповідності. Значення всіх метрик варіантів UD і UD+NER відрізняються в межах 0.01, що вказує на низьку ефективність додаткового застосування поточної моделі пошуку іменованих сутностей. Значення F1-міри для варіанту без використання моделей UD і NER менше 0.1, тобто представлення іменних груп як окремих іменників і особових займенників є малоефективним та недоцільним в задачах, що потребують попереднє виявлення іменних груп в тексті.

В табл. 2 наведено оцінки ефективності роботи розглянутих варіантів застосування моделей для режиму часткової відповідності. Аналогічно до режиму повної відповідності, значення метрик варіантів моделей UD і UD+NER рівні в межах похибки 0.001, що підкреслює низьку ефективність застосування моделі пошуку іменованих сутностей. Значення F1-міри для обох варіантів рівне 0.902, що вказує на доцільність використання варіанту моделі UD для знаходження іменних груп в україномовних текстах.

| Метрика / Варіанти моделей | Точність | Повнота | F-міра |
|---|---|---|---|
| UD+NER | 0.973 | 0.844 | 0.902 |
| UD | 0.974 | 0.843 | 0.902 |
| – | 0.948 | 0.201 | 0.320 |

*Табл. 2 – Оцінка ефективності роботи методу з різними варіантами моделей для режиму часткової відповідності*

**Висновки**

Проаналізовано основні методи пошуку іменних груп та іменованих сутностей для англомовних та україномовних текстів. Методи аналізу англомовного тексту не можуть бути використані для україномовних документів, адже вони створені з урахуванням особливостей структури побудови речень тільки в англомовних текстах. Для виявлення іменованих сутностей в україномовних текстах доцільно використовувати попередньо навчену модель відповідно до предметної області вхідних текстів; можливим є використанням додаткових регулярних виразів для екстракції іменованих сутностей з фіксованою структурою. Проаналізовано результати застосування технології Universal Dependencies для україномовних текстів з метою здійснення перетворення вхідної текстової інформації в деревовидну структуру. На основі аналізу деревовидної структури запропоновано метод пошуку іменованих груп в україномовних документах. Отримано експериментальні результати застосування пропонованого методу з різними варіаціями його використанням: окремо та разом з моделлю пошуку іменованих сутностей в тексті. Розраховані метрики ефективності роботи методу вказують на доцільність його використання для пошуку іменованих груп в україномовних текстах. Для підвищення точності роботи методу можуть бути застосовані наступні підходи:

– використання навченої моделі пошуку іменованих сутностей та набору газетирів відповідно до предметної області;

– застосування набору регулярних виразів для виявлення іменних груп з фіксованою структурою;

– використання сторонніх моделей токенізації тексту для зменшення похибки виявлення частин мови слів тексту.

**Література**

**Погорелый С.Д., Крамов А.А.**

**Метод определения именных групп в украиноязычных текстах**

**Введение**. Отрасль обработки естественного языка рассматривает AI-полные задачи, которые не могут быть решены с помощью алгоритмический действий. Задачи такого типа решаются с использованием методологии машинного обучения и методов компьютерной лингвистики. Одной из задач предварительной обработки текста является поиск именных групп; точность их определения существенно влияет на эффективность решения многих задач обработки естественного языка. Несмотря на активное развитие исследований в направлении обработки естественного языка, исследование поиска именных групп в украиноязычных текстах находится на начальном этапе.

**Цель статьи**. Сравнительный анализ основных методов поиска именных групп в англоязычных и украиноязычных текстах. Создание комплексного метода определения именных групп в текстах соответственно с особенностями украинского языка. Осуществление экспериментальной проверки предложенного метода на корпусе украиноязычных статей.

**Результаты.** Проанализированы методы поиска именных групп в тексте и обоснована целесообразность использования древовидной синтаксической структуры предложения. Недостатком многих методов поиска именных групп в тексте является зависимость эффективности их определения от свойств конкретного языка. Решено использовать модель Universal Dependencies в связи с унифицированным форматом обработки предложения для разных языков и наличием обученной модели построение древовидной структуры предложений украиноязычных текстов. Предложен комплексный метод определения именных групп в украиноязычных текстах с использованием средств Universal Dependencies и модели распознавания именованных сущностей.

Осуществлена экспериментальная проверка эффективности предложенного метода на корпусе украиноязычных новостей и рассчитаны метрики точности метода.

**Выводы**. Полученные результаты рассчитанных метрик точности предложенного метода могут свидетельствовать о целесообразности применение метода для поиска именных групп в украиноязычных текстах. Улучшение точности метода возможно с помощью применения моделей и шаблонов распознавания именованных сущностей в соответствии с рассматриваемой предметной областью.

**Ключевые слова**: обработка естественного языка, именная группа, модель Universal Dependencies, модель NER, древовидная структура предложения.

**Відомості про авторів**

**Погорілий Сергій Дем'янович**

Доктор технічних наук, професор, завідувач кафедри комп'ютерної інженерії факультету радіофізики, електроніки та комп'ютерних систем Київського національного університету імені Тараса Шевченка (Київ, Україна).

Адреса організації: 03022, Київ, проспект Академіка Глушкова, 4Г.

Номер ORCID: https://orcid.org/0000-0002-6497-5056

Електронна пошта: sdp@univ.net.ua

**Крамов Артем Андрійович**

Аспірант кафедри комп'ютерної інженерії факультету радіофізики, електроніки та комп'ютерних систем Київського національного університету імені Тараса Шевченка (Київ, Україна).

Адреса організації: 03022, Київ, проспект Академіка Глушкова, 4Г.

Номер ORCID: https://orcid.org/0000-0003-3631-1268

Електронна пошта: artemkramovphd@knu.ua

**About the authors**

**Pogorilyy Sergiy Demianovych**

Doctor of technical sciences, professor, head of computer engineering department of the faculty of radiophysics, electronics and computer systems of Taras Shevchenko National University of Kyiv (Kyiv, Ukraine).

Address: Glushkov ave., 4G, Kyiv, 03022, Ukraine

**Kramov Artem Andriiovych**

Postgraduate student at the computer engineering department of the faculty of radiophysics, electronics and computer systems of Taras Shevchenko National University of Kyiv (Kyiv, Ukraine).

Address: Glushkov ave., 4G, Kyiv, 03022, Ukraine